\documentclass{article}
\pdfoutput=1

\usepackage{arxiv}

\usepackage{amsmath}

\usepackage{algorithm}
\usepackage{algcompatible}
\usepackage{adjustbox}
\usepackage{amsfonts}
\usepackage{amssymb}
\usepackage{amsbsy}
\usepackage[T1]{fontenc}
\usepackage{textcomp}
\usepackage{graphicx}
\usepackage{listings}
\usepackage{multirow}
\usepackage[normalem]{ulem}
\usepackage{color}
\definecolor{ruby}{rgb}{0.6,0,0.3}
\definecolor{blueice}{rgb}{0.8,0.9,1}

\usepackage{enumitem}
\usepackage{graphicx}
\graphicspath{{./}}

\usepackage[hidelinks]{hyperref} %{url} does not create proper links when '_' is present
\usepackage{fixltx2e}
\usepackage{soul}

\usepackage{tabularx}
\usepackage{longtable}
\usepackage{afterpage} %//flush floating tables without pagebreak to prevent overflow
\usepackage{colortbl}
\newcolumntype{Y}{>{\centering\arraybackslash}X}
\usepackage{tikz}

\newcommand{\ignore}[1]{}
\newcommand{\revisit}[1]{}

\usepackage[utf8]{inputenc} % allow utf-8 input

\usepackage{booktabs}       % professional-quality tables
\usepackage{microtype}      % microtypography
\usepackage{xr-hyper} %//ability to cross-reference external documents given .aux file
\title{Empirical observations on the effects of data transformation in machine learning classification of geological domains}
\shorttitle{Empirical observations: effects of data transformation on machine learning classification}

\author{
  \textbf{Raymond~Leung}\vspace{2mm} \\
  Australian Centre for Field Robotics (ACFR)\\
  Faculty of Engineering\\
  The University of Sydney\\
  Sydney, NSW 2006 \\
  \texttt{raymond.leung@sydney.edu.au} \\
}

\date{}

\renewcommand{\headeright}{}
\renewcommand{\undertitle}{}

\begin{document}
\maketitle

\begin{abstract}In the literature, a large body of work advocates the use of log-ratio transformation for multivariate statistical analysis of compositional data. In contrast, few studies have looked at how data transformation changes the efficacy of machine learning classifiers within geoscience. This letter presents experiment results and empirical observations to further explore this issue. The objective is to study the effects of data transformation on geozone classification performance when machine learning (ML) classifiers\,/\,estimators are trained using geochemical data. The training input consists of exploration hole assay samples obtained from a Pilbara iron-ore deposit in Western Australia, and geozone labels assigned based on stratigraphic units, the absence or presence and type of mineralization.\footnote{In geology, mineralization refers to hydrothermal deposition of economically important metals in the formation of ore bodies.} The ML techniques considered are multinomial logistic regression, Gaussian na\"{i}ve Bayes, kNN, linear support vector classifier, RBF-SVM, gradient boosting and extreme GB, random forest (RF) and multi-layer perceptron (MLP). The transformations examined include isometric log-ratio (ILR), center log-ratio (CLR) coupled with principal component analysis (PCA) or independent component analysis (ICA), and a manifold learning approach based on local linear embedding (LLE). The results reveal that different ML classifiers exhibit varying sensitivity to these transformations, with some clearly more advantageous or deleterious than others. Overall, the best performing candidate is ILR which is unsurprising considering the compositional nature of the data. The performance of pairwise log-ratio (PWLR) transformation is better than ILR for ensemble and tree-based learners such as boosting and RF; but worse for MLP, SVM and other classifiers.
\end{abstract}

\section{Background and Motivation}\label{sec:background-motivation}
In mathematical geoscience literature, log-ratio transformation is generally recommended for multivariate analysis of geochemical or compositional data \cite{kucera1998logratio,verma2006discriminating,verma2015monte,darabi2018evaluation,karacan2018mapping,jaconi2019log,darabi2019multivariate}. Yet, the effects of such transformation on the efficacy of machine learning techniques have received much less attention. Amongst the works that deal with this subject \cite{andrews2015generalized,wang2020comparison,zhang2020systematic,tolosana2019machine}, Tolosana-Delgado et al. \cite{tolosana2019machine} have shown that ensemble tree-based learners, such as random forest\footnote{Random Forest addresses the issue of over-fitting in partition trees by introducing randomness\cite{breiman2001random}. It bootstraps the number of observations by creating different trees that form the forest. Furthermore, branching in each tree is based on a different random subset (rather than all) of the variables.}, benefit more from pairwise log-ratio (PWLR) transformation than CLR or ILR. At the same time, there is no published research on whether similar observations would hold for other classes of learners such as logistic regression \cite{fagerland2008multinomial}, support vector machine \cite{cortes1995support} and artificial neural networks (ANN) \cite{anthony2009neural} which operate on different principles.

The purpose of this paper is to shed light on the role of signal transformation and examine how different transformations affect the geozone classification performance of various machine learning (ML) classifiers. This follows a previous study \cite{leung2021warpml-arxiv} in which ML techniques have been used to estimate the geozone likelihood probability given the sample chemistry, $p(g\!\mid\!\mathbf{c})$. Each chemistry sample $\mathbf{c}\in\mathbb{R}^K$ lies in the Aitchison simplex and satisfies the constraints $c_k\!>\!0$ for all $k$ and $\sum_k c_k=1$. A \textit{geozone} simply refers to a geological domain label that encapsulates knowledge about the stratigraphy and mineralization state \cite{clout2006iron} based on some sequence and processes of geological formation. Naturally, there are chemical characteristics and a spatial structure associated with these geological domains \cite{clout2006iron}. In general, using both chemical and spatial attributes for training can lead to more accurate geozone classification.

In practice, however, exploiting spatial information does not necessarily lead to improved performance. It depends on the application and how the ML predictions are used, so it comes with cautionary notes. Firstly, the exploration hole samples used during training often provide different spatial coverage and might not be compatible with the sampling locations expected at deployment time. Hence, incorporating spatial features during training may lead to over-fitting and unreliable predictions. Secondly, if machine learning emits geozone likelihood estimates rather than a classification, confusion between geozones with similar chemical characteristics may not pose a problem, particularly if a proposal incorporates other mechanisms for resolving such ambiguity. This is the case in \cite{leung2021warpml-arxiv} where a spatial proximity filter essentially eliminates infeasible geozone candidates that are chemically similar but physically distant from a test sample. This was done in the context of a complex system where ML probabilistic estimates were used in conjunction with a surface warping algorithm \cite{leung2020bayesian} to rectify spatial inaccuracies in the boundaries that delineate various geozones.

As a follow-up study to \cite{leung2021warpml-arxiv}, this paper focuses exclusively on the task of estimating $p(g\!\mid\!f(\mathbf{c}))$ using ML, where $f(\mathbf{c})$ denotes some data transformation such as isometric log-ratio (ILR), or center log-ratio (CLR) coupled with principal component analysis (PCA), independent component analysis (ICA) or local linear embedding (LLE). When $f$ equals the identity mapping, there is no data transformation and the original geochemical assay data is used as is. The geochemical dataset contains roughly 35,000 assay samples which is split 60:40 for training and testing. Each sample provides a measurement of the concentration of Fe, SiO\textsubscript{2}, Al\textsubscript{2}O\textsubscript{3}, P etc.\,at a different location.

For evaluation, two scenarios are of interest. The first scenario involves a very ambitious task where (i) the distinction between 46 individual geozones are maintained and (ii) the ML estimators have to emit a likelihood estimate for each of the geozones \cite{zadrozny2002transforming}. Since some samples between certain geozones are chemically indistinguishable, identifying \textit{the} correct geozone will be exceedingly difficult. Such difficulties had previously been highlighted by examining the silhouette score \cite{rousseeuw-87}, confusion matrix and estimates for the conditional probabilities $\hat{p}(g_\text{predict}\!\mid\!\mathbf{c},g_\text{actual})$ in Section~\ignore{\ref{sec:ml-classifier-analysis}}4.1 of \cite{leung2021warpml-arxiv}. This test case is designed to push the ML candidates to their limit. The typical geochemical composition of the iron ore test data used in these experiments is described in \cite{clout2006iron} and \cite{leung2019sample}.

Precision and recall (also $F_1$ score) are reported, along with the top-$n$ recognition rates. The top-$n$ recognition rate reports the percentage of samples for which the true class label appears in the $n$ most likely geozones. In the second scenario, the problem is simplified by aggregating the probabilities for geozones that belong to the same group. Using set notation, $M$, $H$ and $U$ denote mineralized, hydrated, and unmineralized domains. So, for instance, $p(g=M_i)$ and $p(g=M_j)$ are combined. The geozone likelihood question is then posed in a binary context, it asks whether $p(g\in M\cup H)$ is more likely than $p(g\in U)$. In propositional logic, $(\text{M}\vee\text{H}) \Rightarrow\neg\text{U}$.

The ML classifiers\,/\,estimators considered in these experiments cover a variety of well known algorithms implemented in the scikit-learn library (v0.22.1) \cite{pedregosa2011scikit}, they include:
\begin{itemize}
\item Multinomial logistic regression (Logistic) \cite{yu2011dual}\cite{zhu1997algorithm}
\item Gaussian na\"{i}ve Bayes (GaussianNB) \cite{murphy2006naive}
\item k-nearest neighbors (KNN) \cite{song2017efficient}
\item Linear support vector classifier (L-SVC) \cite{chang2011libsvm}
\item Radial basis function support vector machine (RBF-SVM) \cite{cortes1995support}
\item Gradient boosting (GradBoost) \cite{friedman2001greedy}
\item Extreme gradient boosting (XGB) \cite{chen2016xgboost}
\item Random forest (RF) \cite{breiman2001random}
\item Mult-layer perceptron (MLP) \cite{hinton1990connectionist}\cite{glorot2010understanding}\cite{he2015delving}\cite{kingma2014adam} (a feedforward neural network with fully-connected layers as configured in \cite{leung2021warpml-arxiv})
\end{itemize}

The objective is to provide anecdotal evidence on the effects of data transformation for a classification task, specifically in identifying geozones (geological domains) using machine learning and geochemical data gathered from an iron ore mine. Through sharing these results, it is hoped the test scenarios can provide some guidance for similar problems that appear at the junction of machine learning and geoscience in the mineral resources sector; particularly on the appropriateness of data transformations and robustness of different ML classifiers.

\section{Baseline configuration: geochemistry without data transformation}\label{sec:supp-baseline}
The baseline performance is established in Table~\ref{supp-tab:multi-class-geozone-ml-performance}. This corresponds to the case $f\equiv\mathcal{I}$ where no data transformation has been applied to the raw geochemical data, $\mathbf{c}$. As noted in \cite{leung2021warpml-arxiv}, the RGB-SVM and KNN both struggle to achieve an $F_1$ score anywhere close to RF. Both methods measure differences using the squared Euclidean distance, yet feature vectors that are the same distance apart in the Aitchison simplex
\begin{align}
\mathbf{c}\in\mathcal{S}^K=\left\{\left.\mathbf{c}=[c_1,c_2,\ldots,c_K]\in\mathbb{R}^K\right| c_k>0,k=1,2,\ldots,K;\,\sum_{k=1}^K c_k=1\right\}\label{eq:simplex-constraints}
\end{align}
may not be equally dissimilar because the data has to satisfy the compositional constraints in (\ref{eq:simplex-constraints}).\footnote{Here, $c_1,\ldots,c_K$ are used to represent the concentration of Fe, SiO\textsubscript{2}, Al\textsubscript{2}O\textsubscript{3}, P, LOI (loss on ignition), TiO\textsubscript{2}, MgO, Mn, CaO and S measured in an assay sample.} The geochemical significance of these differences depend on the components in which these differences occur. As the training data is not symmetrically distributed, the interpretation of margins and structural risk minimization may also be affected.

\begin{table}[!ht]
\begin{center}
\small
\setlength\tabcolsep{4pt}
\caption{Geozone classification performance for various machine learning techniques\\ (reproduced from Tables~\ignore{\ref{supp-tab:multi-class-geozone-ml-performance}}1 and \ignore{\ref{tab:multi-class-geozone-top-n-recall-rates}}3 in \cite{leung2021warpml-arxiv})}\label{supp-tab:multi-class-geozone-ml-performance}
\resizebox{\textwidth}{!}{
\begin{tabular}{|l|cccc|cccc||cccc|}\hline
&\multicolumn{4}{c|}{46 geozones}&\multicolumn{4}{c||}{Top-$n$ recognition rate (\%)}&\multicolumn{4}{c|}{Aggregated geozones (M$\vee$H) vs U}\\ \hline
Classifier & Brier & Precision & Recall & $F_1$ score & $n=1$ & $n=2$ & $n=4$ & $n=8$ & Brier & Precision & Recall & $F_1$ score \\ \hline
Logistic & 0.6603 & 45.57 & 51.41 & 48.31 & 51.40 & 71.92 & 84.61 & 94.40 & 0.0512 & 94.01 & 94.02 & 94.02\\
GaussianNB & 0.6531 & 46.41 & 51.78 & 48.95 & 51.78 & 72.18 & 86.03 & 94.71 & 0.0461 & 94.10 & 94.05 & 94.08\\
KNN & 0.6266 & 48.57 & 53.45 & 50.89 & 53.45 & 71.99 & 82.33 & 88.17 & 0.0542 & 94.39 & 94.40 & 94.40\\
L-SVC & 0.5799 & 52.18 & 57.45 & 54.69 & 57.45 & 76.41 & 89.75 & 96.51 & 0.0381 & 94.97 & 94.95 & 94.96\\
RBF-SVM & 0.6174 & 48.28 & 53.95 & 50.95 & 53.95 & 74.31 & 87.51 & 95.73 & 0.0389 & 95.01 & 94.99 & 95.00\\
GradBoost & 0.4965 & 61.63 & 64.12 & 62.85 & 64.11 & 81.58 & 92.18 & 97.14 & 0.0382 & 95.00 & 94.99 & 95.00\\
XGB & 0.4721 & 63.60 & 65.79 & 64.68 & 65.79 & 83.07 & 93.31 & 98.00 & 0.0356 & 95.26 & 95.26 & 95.26\\
MLP & 0.4645 & 64.24 & 66.31 & 65.25 & 66.30 & 83.59 & 93.67 & 98.07 & 0.0361 & 95.16 & 95.16 & 95.16\\
RF & 0.4319 & 67.95 & 68.98 & 68.46 & 68.98 & 85.56 & 94.60 & 98.22 & 0.0341 & 95.43 & 95.42 & 95.43\\ \hline
\end{tabular}
}
\end{center}
\end{table}

\section{Variant 1: geochemistry with isometric log-ratio transformation (ILR)}\label{sec:supp-with-ilr}
Isometric log-ratio transformation \cite{egozcue-03} is a standard tool for analyzing compositional data. The ILR isometry is obtained by taking a center log-ratio (CLR) transformation, followed by Gram-Schmidt orthogonalization. The orthogonal matrix may be computed by constructing log-contrasts using a bifurcating tree (sequential binary partitions). For this work, an alternative construction that inherits from the Helmert sub-matrix (described by Tsagris et al.\,\cite{tsagris-2011-data}) is adopted.

Mathematically, the ILR is often regarded as the orthodox approach for opening up closed data whereby the components in the raw geochemical measurements, $c_k$, are initially subject to the \textit{non-negative value} and \textit{constant sum} constraints \cite{kucera1998logratio}. The data in its original form (before ILR transformation) are far from symmetric and certainly not Gaussian distributed. Qualitative differences between the raw data and ILR transformed data can be seen in Fig.~4 in \cite{leung2019sample}. Without ILR, the data in this ``constricted'' space may not meet the expectation of certain ML classifiers. For instance, in terms of data spread and pairwise affinity, the compositional scale and covariance may create locally varying (non-stationary) anisotropic conditions \cite{boisvert2009calculating}. Depending on the loss functions used, some classifiers might not fully recognize similarities in the simplicial sample space (in the way it was intended), or penalise differences unfairly in goodness-of-fit evaluation. Thus, it is not unreasonable to think the results presented in Table~\ref{supp-tab:multi-class-geozone-ml-performance} might be sub-optimal.

This indeed is evident from the results presented in Table~\ref{supp-tab:multi-class-geozone-ilr-performance}. Looking at the changes relative to the baseline configuration (see $\Delta F_1$ column), performance has improved for all ML estimators except gradient boost. As discussed in \cite{leung2021warpml-arxiv}, KNN and RBF-SVM benefit the most from ILR transformation. The k nearest neighbor algorithm and radial basis function in RBF-SVM are inherently sensitive to asymmetric distortion in the data which is substantially corrected by the ILR. Also worth noting is that RF is impacted the least while MLP and RF both reach new heights, with an $F_1$ score of 69.5\% and 68.7\%,  respectively. The top-$n$ recognition rates also improve in a similar manner.

\begin{table}[!ht]
\small
\setlength\tabcolsep{5pt}
\caption{Geozone classification performance for isometric log-ratio transformed chemical data\\ (reproduced from Table~\ignore{\ref{tab:multi-class-geozone-ilr-performance}}6 in \cite{leung2021warpml-arxiv})}\label{supp-tab:multi-class-geozone-ilr-performance}
\resizebox{\textwidth}{!}{
\begin{tabular}{|l|cccc|cc|cccc||cc|}\hline
&\multicolumn{4}{c|}{46 geozones} & \multicolumn{2}{c|}{Change (rel. Table~\ref{supp-tab:multi-class-geozone-ml-performance})} &\multicolumn{4}{c||}{Top-$n$ recognition rate (\%)} &\multicolumn{2}{c|}{(M$\vee$H) vs U}\\ \hline
Classifier & Brier & Precision & Recall & $F_1$ score & $\Delta$ Brier & $\Delta F_1$ & $n=1$ & $n=2$ & $n=4$ & $n=8$ & Brier & $F_1$ score\\ \hline
Logistic & 0.5849 & 52.02 & 57.86 & 54.79 & -0.0754 & +6.48 & 57.86 & 75.50 & 87.57 & 95.62 & 0.0463 & 94.26\\
GaussianNB & 0.5728 & 54.64 & 57.75 & 56.15 & -0.0803 & +7.20 & 57.75 & 75.70 & 88.32 & 96.03 & 0.0459 & 93.89\\
KNN & 0.4389 & 67.34 & 68.61 & 67.97 & -0.1339 & +17.08\textsuperscript{*} & 68.61 & 85.13 & 92.37 & 94.90 & 0.0412 & 94.82\\
L-SVC & 0.5289 & 56.50 & 61.20 & 58.76 & -0.0510 & +4.07 & 61.20 & 79.82 & 91.69 & 97.64  & 0.0378 & 94.96\\
RBF-SVM & 0.4993 & 61.02 & 63.63 & 62.30 & -0.1181 & +11.35\textsuperscript{*} & 63.63 & 81.91 & 93.09 & 98.18 & 0.0386 & 94.85\\
GradBoost & 0.5162 & 60.57 & 63.16 & 61.84 & +0.0197 & \textbf{\color{ruby}-1.01} & 63.16 & 81.12 & 91.61 & 96.56 & 0.0413 & 94.58\\
XGB & 0.4719 & 63.54 & 65.68 & 64.59 & -0.0002 & -0.05 & 65.68 & 83.20 & 93.44 & 98.06 & 0.0356 & 95.20\\
MLP & 0.4198 & 68.92 & 70.13 & \textbf{69.52} & -0.0447 & \textbf{+4.27} & 70.13 & 86.43 & 95.17 & 98.63 & 0.0346 & 95.47\\
RF & 0.4281 & 68.28 & 69.23 & \textbf{68.75} & -0.0038 & \textbf{+0.29} & 69.23 & 85.86 & 94.82 & 98.32 & 0.0339 & 95.47\\\hline
\multicolumn{11}{c}{A positive $\Delta F_1$ score (likewise a negative $\Delta$Brier score) indicates performance gain relative to $p(g\!\mid\!\mathbf{c})$ from Table~\ref{supp-tab:multi-class-geozone-ml-performance}}\\
\end{tabular}
}
\end{table}

\newpage
\section{Variant 2: geochemistry with principal component analysis (PCA) and whitening}\label{sec:supp-with-pca-whiten}
Principal component analysis (PCA) is arguably the best known multivariate technique for extracting information and analyzing structure amongst correlated variables. This decorrelating transform projects data onto a set of orthonormal bases and has the optimal property that it achieves the highest energy compaction amongst linear transformations. In other words, for a given number of leading principal components (say $m$), the $m$ PCA coefficients are able to explain the variance observed in the data to the largest possible extent. This is achieved by aligning the prototype vectors (or principal components) with the axes of the $m$-dimensional ellipsoid fitted by the data. The underlying rotation allows the data to be portrayed in the right frame where meaningful changes in the signal become prominent.

Although PCA is often used for dimensionality reduction or to suppress noise, this turns out to be unhelpful for our classification problem --- it discards too much information against the backdrop of significant inter-class overlap and requirement of high specificity. There are numerous geozones (classes) present in the annotated data which makes the task very challenging. For this reason, the full dimensions of the input are retained and none of the 10 transformed coefficients are truncated in all subsequent experiments. Furthermore, whitening is performed.\footnote{The component vectors are multiplied by the square root of the number of samples, $\sqrt{N}$, and then divided by the singular values to ensure the output components have unit variance.} The projection matrix can be obtained using eigen-decomposition and singular value decomposition (SVD) given an $N$ sample observation matrix $X\in\mathbb{R}^{N\times K}$. The details are described in \cite{abdi2010principal}.

The first set of experiments apply PCA directly to the geochemical measurements. The results in Table~\ref{supp-tab:multi-class-geozone-pca-whiten-performance} serve to demonstrate that this is not the right way of handling compositional data.
\begin{table}[!ht]
\small
\setlength\tabcolsep{5pt}
\caption{Geozone classification performance for PCA transformed chemical data with whitening}\label{supp-tab:multi-class-geozone-pca-whiten-performance}
\resizebox{1.0\textwidth}{!}{
\begin{tabular}{|l|cccc|cc|cccc||cc|}\hline
&\multicolumn{4}{c|}{Multi-class (46 unique geozones)} & \multicolumn{2}{c|}{Change (rel. Table~\ref{supp-tab:multi-class-geozone-ml-performance})} &\multicolumn{4}{c||}{Top-$n$ recognition rate (\%)} &\multicolumn{2}{c|}{(M$\vee$H) vs U}\\ \hline
Classifier & Brier & Precision & Recall & $F_1$ score & $\Delta$ Brier & $\Delta F_1$ & $n=1$ & $n=2$ & $n=4$ & $n=8$ & Brier & $F_1$ score\\ \hline
Logistic & 0.5874 & 52.62 & 57.40 & 54.90 & -0.0729 & +6.60 & 57.40 & 76.31 & 88.69 & 95.86 & 0.0390 & 94.75\\
GaussianNB & 0.6818 & 44.32 & 48.77 & 46.44 & +0.0286 & \textbf{\color{ruby}-2.51} & 48.77 & 70.29 & 84.32 & 93.50 &  0.0477 & 94.04\\
KNN & 0.5218 & 59.98 & 62.25 & 61.10 & -0.1049 & +10.20 & 62.25 & 79.18 & 88.18 & 91.98  & 0.0473 & 94.39\\
L-SVC & 0.5640 & 53.61 & 58.75 & 56.06 & -0.0159 & +1.37 & 58.75 & 77.98 & 90.71 & 97.03 & 0.0381 & 95.01\\
RBF-SVM & 0.5288 & 58.47 & 61.45 & 59.92 & -0.0866 & +8.96 & 61.45 & 79.64 & 91.34 & 97.29 & 0.0387 & 94.89\\
GradBoost & 0.5593 & 58.13 & 60.72 & 59.40 & +0.0628 & \textbf{\color{ruby}-3.45} & 60.72 & 79.04 & 90.00 & 95.08 & 0.0458 & 94.17\\
XGB & 0.4933 & 61.59 & 64.00 & 62.77 & +0.0213 & \textbf{\color{ruby}-2.01} & 64.01 & 81.61 & 92.36 & 97.53 & 0.0368 & 95.06\\
MLP & 0.4468 & 65.95 & 67.68 & 66.81 & -0.0178 & +1.55 & 67.68 & 84.65 & 94.33 & 98.33 & 0.0361 & 95.17\\
RF & 0.4632 & 65.03 & 66.58 & 65.80 & +0.0313 & \textbf{\color{ruby}-2.67} & 66.58 & 83.29 & 93.25 & 97.63 & 0.0360 & 95.28\\ \hline
\multicolumn{11}{c}{A positive $\Delta F_1$ score (likewise a negative $\Delta$Brier score) indicates performance gain relative to $p(g\!\mid\!\mathbf{c})$ from Table~\ref{supp-tab:multi-class-geozone-ml-performance}}\\
\end{tabular}
}
\end{table}
\begin{itemize}
\item When the PCA decorrelating transform is applied directly to the raw geochemical data, the performance deteriorates significantly for all classifiers except logistic regression which changes very little.
\item In fact, the $F_1$ scores have dropped relative to ILR almost throughout. For L-SVC, RBF-SVM, GradBoost, MLP and RF, the change in $\Delta F_1$ (gap between ILR and PCA) ranges from -2.39\% to -2.96\%. For GaussianNB and KNN, this gap grows to -9.7\% and -6.88\%, respectively.
\item Even when an improvement is observed relative to the baseline (see positive $\Delta F_1$ for RGB-SVM and MLP in Table~\ref{supp-tab:multi-class-geozone-pca-whiten-performance}), the performance gain is significantly smaller compared with the same figures in Table~\ref{supp-tab:multi-class-geozone-ilr-performance}.
\item GaussianNB, GradBoost and RF all perform worse than the baseline.
\item These results show conclusively that applying PCA directly to the raw geochemical data attracts a heavy performance penalty when ML geozone classification is attempted using the PCA transformed data.
\end{itemize}

\section{Variant 3: geochemistry with centered log-ratio transformation (CLR) followed by principal component analysis (PCA)}\label{sec:supp-with-clr-pca}
The second set of experiments apply a CLR transformation \cite{egozcue-03} to the geochemical measurements before the PCA. The results in Table~\ref{supp-tab:multi-class-geozone-clr-pca-performance} demonstrate a better way of handling compositional data.
\begin{itemize}
\item CLR is an isomorphism and isometry that transforms the Aitchison simplex to real space.
\item It ``opens up'' the compositional data and makes PCA more effective than without a log-ratio transform.
\item KNN and RBF-SVM are highly sensitive to noise in the data and their performance improve with CLR+PCA (though not as much as from ILR) while MLP also benefits.
\item The performance of ensemble classifiers such as GradBoost, XGB and RF are degraded when PCA is applied. This may be due to the energy compaction achieved by SVD which results in a high concentration of the variance explained by the leading principal components. The resultant PCA coefficients do not produce features with uniform significance, thus it may lead to more uneven splits during the regression/decision tree building process.
\end{itemize}

\begin{table}[!ht]
\small
\setlength\tabcolsep{5pt}
\caption{Geozone classification performance for CLR-PCA transformed chemical data}\label{supp-tab:multi-class-geozone-clr-pca-performance}
\resizebox{1.0\textwidth}{!}{
\begin{tabular}{|l|cccc|cc|cccc||cc|}\hline
&\multicolumn{4}{c|}{Multi-class (46 unique geozones)} & \multicolumn{2}{c|}{Change (rel. Table~\ref{supp-tab:multi-class-geozone-ml-performance})} &\multicolumn{4}{c||}{Top-$n$ recognition rate (\%)} &\multicolumn{2}{c|}{(M$\vee$H) vs U}\\ \hline
Classifier & Brier & Precision & Recall & $F_1$ score & $\Delta$ Brier & $\Delta F_1$ & $n=1$ & $n=2$ & $n=4$ & $n=8$ & Brier & $F_1$ score\\ \hline
Logistic & 0.5517 & 54.49 & 59.46 & 56.86 & -0.1086 & +8.56 & 59.46 & 77.65 & 89.50 & 96.29 & 0.0405 & 94.53\\
GaussianNB & 0.5821 & 53.43 & 56.88 & 55.10 & -0.0710 & +6.15 & 56.88 & 74.92 & 87.52 & 95.18 &  0.0521 & 92.99\\
KNN & 0.4401 & 67.30 & 68.54 & 67.91 & -0.1865 & +17.01 & 68.54 & 85.02 & 92.32 & 94.88 & 0.0415 & 94.79\\
L-SVC & 0.5340 & 55.86 & 60.73 & 58.19 & -0.0459 & +3.50 & 60.73 & 79.49 & 91.49 & 97.52 & 0.0386 & 94.90\\
RBF-SVM & 0.4802 & 62.92 & 65.09 & 63.98 & -0.1373 & +13.03 & 65.09 & 83.31 & 93.92 & 98.40 & 0.0385 & 94.90\\
GradBoost & 0.5601 & 57.96 & 60.63 & 59.26 & +0.0635 & \textbf{\color{ruby}-3.59} & 60.63 & 79.16 & 89.81 & 94.82 & 0.0546 & 93.00\\
XGB & 0.4890 & 61.95 & 64.36 & 63.13 & +0.0170 & \textbf{\color{ruby}-1.54} & 64.36 & 82.04 & 92.59 & 97.70 & 0.0414 & 94.31\\
MLP & 0.4173 & 69.32 & 70.42 & \textbf{69.86} & -0.0473 & \textbf{+4.61} & 70.42 & 86.66 & 95.18 & 98.67 & 0.0354 & 95.39\\
RF & 0.4351 & 67.71 & 68.72 & 68.21 & +0.0032 & \textbf{\color{ruby}-0.25} & 68.72 & 85.60 & 94.63 & 98.30 & 0.0386 & 94.78\\ \hline
\multicolumn{11}{c}{A positive $\Delta F_1$ score (likewise a negative $\Delta$Brier score) indicates performance gain relative to $p(g\!\mid\!\mathbf{c})$ from Table~\ref{supp-tab:multi-class-geozone-ml-performance}}\\
\end{tabular}
}
\end{table}
Overall, the $\Delta F_1$ values reported in Table~\ref{supp-tab:multi-class-geozone-clr-pca-performance} (CLR+PCA) are similar to those in Table~\ref{supp-tab:multi-class-geozone-ilr-performance} (for ILR) --- see KNN (+17.01 vs +17.08) and MLP (+4.61 vs +4.27) for instance. Theoretically, these approaches belong to two equivalent classes. The key difference being in the case of CLR+PCA, the orthogonal bases are obtained by applying singular-value decomposition to CLR transformed data.

\section{Variant 4: geochemistry with centered log-ratio transformation (CLR) followed by independent component analysis (ICA)}\label{sec:supp-with-clr-ica}
While the goal in PCA is to find an orthogonal linear transformation that maximizes the variance of the variables, the goal of ICA is to discover fundamental structures within the data and decompose the signal into statistically independent and non-Gaussian components \cite{hyvarinen2000independent}. It is frequently used in source separation (unmixing) problems. A key difference to PCA is the resultant basis vectors are not required to be orthogonal or rank ordered.

The main observation from Table~\ref{supp-tab:multi-class-geozone-clr-ica-performance} is that ICA offers no advantage compared with CLR+PCA in the ML geozone classification problem. The ICA transformed variables appear to be non-linearly separable, as evident from the $\Delta F_1$ values: -4.27 for L-SVC and +14.14 for RBF-SVM which represent a change of -7.77 and +1.11, respectively, relative to CLR+PCA. Interestingly, the $F_1$ score of the leading classifier MLP also drops by -1.66 relative to the baseline.

\begin{table}[!ht]
\small
\setlength\tabcolsep{5pt}
\caption{Geozone classification performance for CLR-ICA transformed chemical data (with standardized components)}\label{supp-tab:multi-class-geozone-clr-ica-performance}
\resizebox{1.0\textwidth}{!}{
\begin{tabular}{|l|cccc|cc|cccc||cc|}\hline
&\multicolumn{4}{c|}{Multi-class (46 unique geozones)} & \multicolumn{2}{c|}{Change (rel. Table~\ref{supp-tab:multi-class-geozone-ml-performance})} &\multicolumn{4}{c||}{Top-$n$ recognition rate (\%)} &\multicolumn{2}{c|}{(M$\vee$H) vs U}\\ \hline
Classifier & Brier & Precision & Recall & $F_1$ score & $\Delta$ Brier & $\Delta F_1$ & $n=1$ & $n=2$ & $n=4$ & $n=8$ & Brier & $F_1$ score\\ \hline
Logistic & 0.6165 & 47.31 & 53.93 & 50.40 & -0.0438 & +2.10 & 53.93 & 73.03 & 86.78 & 94.84 & 0.0525 & 93.45\\
GaussianNB & 0.5807 & 53.33 & 56.85 & 55.04 & -0.0724 & +6.09 & 56.86 & 74.83 & 87.46 & 95.31 &  0.0501 & 93.24\\
KNN & 0.4365 & 67.74 & 68.89 & 68.31 & -0.1901 & +17.41 & 68.89 & 84.98 & 92.40 & 95.12 & 0.0425 & 94.63\\
L-SVC & 0.6375 & 47.30 & 54.24 & 50.54 & +0.0474 & \textbf{\color{ruby}-4.27} & 54.24 & 72.64 & 85.58 & 94.43 & 0.0593 & 93.38\\
RBF-SVM & 0.4688 & 64.02 & 66.22 & 65.10 & -0.1487 & +14.14 & 66.22 & 84.00 & 94.33 & 98.55 & 0.0376 & 95.00\\
GradBoost & 0.5557 & 58.62 & 61.08 & 59.82 & +0.0591 & \textbf{\color{ruby}-3.03} & 61.08 & 79.23 & 89.83 & 94.85 & 0.0522 & 93.42\\
XGB & 0.4827 & 62.60 & 64.86 & 63.71 & +0.0106  & \textbf{\color{ruby}-0.97} & 64.86 & 82.44 & 92.78 & 97.85 & 0.0386 & 94.76\\
MLP & 0.4813 & 62.24 & 65.00 & 63.59 & +0.0167 & \textbf{\color{ruby}-1.66} & 65.00 & 82.54 & 93.05 & 97.91 & 0.0362 & 95.10\\
RF & 0.4365 & 67.68 & 68.64 & 68.16 & +0.0046 & \textbf{\color{ruby}-0.31} & 68.64 & 85.48 & 94.55 & 98.28 & 0.0373 & 95.06\\ \hline
\multicolumn{11}{c}{A positive $\Delta F_1$ score (likewise a negative $\Delta$Brier score) indicates performance gain relative to $p(g\!\mid\!\mathbf{c})$ from Table~\ref{supp-tab:multi-class-geozone-ml-performance}}\\
\end{tabular}
}
\end{table}

Note: Empirically, the performance upper bound observed for the 46 geozone classification problem is approximately 70\% as measured by the $F_1$ score. As mentioned in \cite{leung2021warpml-arxiv}, when a distinction between individual geozones is maintained, there is considerable overlap between certain geozones (classes) that belong to similar groups. Not every problem can be solved with 90\%+ accuracy. In this geozone classification problem, the chemical composition for certain samples drawn from similar geozones (for instance, $M_i$ and $M_j$ where both belong to mineralized domains) may be completely indistinguishable. For the geological domain likelihood estimation problem posed in \cite{leung2021warpml-arxiv}, confusion between $M_i$ and $M_j$ do not have a material effect on surface warping performance. Hence, the aggregated geozone probability, defined as the probability of a geozone belonging to group $M\cup H$ vs group $U$, provides a more realistic measure of their performance; here, the $F_1$ score varies from 93.2\% to 95.1\%. The consistency of the ML estimators may be inferred from the top-$n$ recognition rate. For instance, the geozones proposed by KNN has a recall rate that saturates at 95.1\% (the hit-rate based on the top 8 candidates) whereas the more potent RF classifier reaches 98.2\%.

\section{Variant 5: geochemistry with centered log-ratio transformation (CLR) followed by linear local embedding (LLE)}\label{sec:supp-with-clr-lle}
Local linear embedding is an unsupervised algorithm proposed by Saul and Roweis \cite{saul2003think}.  It may be viewed as computing a series of local PCA which are globally compared to find the best nonlinear embedding. It computes low dimensional, neighborhood preserving embeddings of high dimensional data by solving a sparse eigenvalue problem. Our exploratory analysis has shown that the geozone classification problem is not helped by projecting the data to lower dimensions (e.g. from $\mathbb{R}^{10}$ to $\mathbb{R}^{6}$) via LLE and performance is maximized with a neighborhood size of 128. The main observation from Table~\ref{supp-tab:multi-class-geozone-clr-lle-performance} is that the best performance achieved under CLR-LLE is consistently inferior to ILR, CLR-PCA (even CLR-ICA) for all classifiers. Hence, LLE does not appear to be suitable for the problem on hand.

\begin{table}[!ht]
\small
\setlength\tabcolsep{5pt}
\caption{Geozone classification performance for CLR-LLE transformed chemical data (with standardized components)}\label{supp-tab:multi-class-geozone-clr-lle-performance}
\resizebox{1.0\textwidth}{!}{
\begin{tabular}{|l|cccc|cc|cccc||cc|}\hline
&\multicolumn{4}{c|}{Multi-class (46 unique geozones)} & \multicolumn{2}{c|}{Change (rel. Table~\ref{supp-tab:multi-class-geozone-ml-performance})} &\multicolumn{4}{c||}{Top-$n$ recognition rate (\%)} &\multicolumn{2}{c|}{(M$\vee$H) vs U}\\ \hline
Classifier & Brier & Precision & Recall & $F_1$ score & $\Delta$ Brier & $\Delta F_1$ & $n=1$ & $n=2$ & $n=4$ & $n=8$ & Brier & $F_1$ score\\ \hline
Logistic & 0.6781 & 44.45 & 52.71 & 48.23 & +0.0178 & -0.08 & 52.71 & 70.38 & 82.24 & 91.72 & 0.0762 & 91.95\\
GaussianNB & 0.6059 & 51.62 & 55.06 & 53.29 & -0.0473 & +4.33 & 55.06 & 73.86 & 86.46 & 94.68 &  0.0568 & 92.85\\
KNN & 0.4269 & 65.31 & 66.79 & 66.04 & -0.1638 & +15.14 & 66.79 & 83.67 & 91.45 & 94.35 & 0.0443 & 94.49\\
L-SVC & 0.6516 & 44.17 & 51.38 & 47.50 & +0.0616 & \textbf{\color{ruby}-7.31} & 51.38 & 71.85 & 85.34 & 94.30 & 0.0609 & 92.71\\
RBF-SVM & 0.4925 & 61.81 & 64.01 & 62.89 & -0.1250 & +11.93 & 64.01 & 82.34 & 93.33 & 98.22 & 0.0395 & 94.69\\
GradBoost & 0.5676 & 57.09 & 59.80 & 58.41 & +0.0711 & \textbf{\color{ruby}-4.44} & 59.80 & 78.76 & 89.79 & 94.88 & 0.0545 & 92.89\\
XGB & 0.4988 & 60.84 & 63.22 & 62.01 & +0.0268 & \textbf{\color{ruby}-2.67} & 63.23 & 81.52 & 92.57 & 97.56 & 0.0424 & 94.17\\
MLP & 0.4976 & 60.49 & 63.44 & 61.93 & +0.0330 & \textbf{\color{ruby}-3.33} & 63.44 & 81.36 & 92.23 & 97.56 & 0.0373 & 94.85\\
RF & 0.4569 & 65.69 & 66.94 & 66.31 & +0.0250 & \textbf{\color{ruby}-2.15} & 66.94 & 84.51 & 94.01 & 97.98 & 0.0420 & 94.27\\ \hline
\multicolumn{11}{c}{A positive $\Delta F_1$ score (likewise a negative $\Delta$Brier score) indicates performance gain relative to $p(g\!\mid\!\mathbf{c})$ from Table~\ref{supp-tab:multi-class-geozone-ml-performance}}\\
\end{tabular}
}
\end{table}

\section{Variant 6: geochemistry with pairwise log-ratio transformation (PWLR)}\label{sec:supp-with-pwlr}
Pairwise log-ratio (PWLR) transformation has been shown to work well with tree-based classifiers. In \cite{tolosana2019machine}, Tolosana-Delgado et al.\,argued that the single-variable branching decisions within a Random Forest can account for the compositional nature of the data more effectively when features are constructed using PWLR. It is well known that compositional differences between samples from different classes are more accurately reflected through relative (rather than absolute) concentration of the constituting components \cite{pawlowsky2006compositional}. Pairwise log-ratios is perhaps the simplest and most intuitive way for capturing such constrast. The PWLR transformation is given by
\begin{align}
\text{pwlr}(\mathbf{c})=\left[\,\xi_{ij}\mid i<j=1,2,\ldots,K\right]\quad\text{where }\xi_{ij}=\text{ln}(c_i/c_j)
\end{align}
which lends itself to Aitchison's notion of compositional distance between two vectors $\mathbf{c}^{(p)}$ and $\mathbf{c}^{(q)}$,
\begin{align}
d^2_A(\mathbf{c}^{(p)},\mathbf{c}^{(q)})=\frac{1}{K^2}\sum^{K}_{i<j}\left(\text{ln}\frac{c^{(p)}_i}{c^{(p)}_j}-\text{ln}\frac{c^{(q)}_i}{c^{(q)}_j}\right)^2
\end{align}
In practice, Laplace smoothing is applied and each pairwise log-ratio is computed as $\xi_{ij}=\text{ln}((c_i+\alpha)/(c_j+K\alpha))$, where $\alpha$ is a small positive constant, for instance, $\alpha= 1\times 10^{-7}$.

\begin{table}[!ht]
\small
\setlength\tabcolsep{3pt}
\caption{Geozone classification performance for pairwise log-ratio transformed chemical data}\label{supp-tab:multi-class-geozone-pwlr-performance}
\resizebox{1.0\textwidth}{!}{
\begin{tabular}{|l|cccc|cc|cccc||cc|}\hline
&\multicolumn{4}{c|}{46 geozones} & \multicolumn{2}{c|}{Change (rel. Table~\ref{supp-tab:multi-class-geozone-ilr-performance})} &\multicolumn{4}{c||}{Top-$n$ recognition rate (\%)} &\multicolumn{2}{c|}{(M$\vee$H) vs U}\\ \hline
Classifier & Brier & Precision & Recall & $F_1$ score & $\Delta$ Brier & $\Delta F_1$ & $n=1$ & $n=2$ & $n=4$ & $n=8$ & Brier & $F_1$ score\\ \hline
Logistic & 0.5673 & 53.08 & 58.49 & 55.65 & -0.0176 & +0.87 & 58.49 & 76.02 & 88.32 & 95.54 & 0.0428 & 94.21\\
GaussianNB & 0.5905 & 53.95 & 56.20 & 55.05 & +0.0177 & \textbf{\color{ruby}-1.10} & 56.20 & 74.05 & 86.80 & 94.80 & 0.0513 & 93.44\\
KNN & 0.4409 & 67.18 & 68.43 & 67.80 & +0.0020 & {\color{ruby}-0.17} & 68.43 & 85.01 & 92.37 & 94.97 & 0.0413 & 94.80\\
L-SVC & 0.5345 & 55.64 & 60.63 & 58.03 & +0.0056 & \textbf{\color{ruby}-0.73} & 60.63 & 79.46 & 91.58 & 97.56 & 0.0387 & 94.89\\
RBF-SVM & 0.5189 & 58.81 & 62.07 & 60.40 & +0.0196 & \textbf{\color{ruby}-1.90} & 62.07 & 80.48 & 92.06 & 97.74 & 0.0409 & 94.57\\
GradBoost & 0.5314 & 61.74 & 63.26 & 62.49 & +0.0153 & +0.65 & 63.26 & 81.13 & 91.01 & 95.25 & 0.0463 & 94.15\\
XGB & 0.4354 & 68.38 & 69.40 & 68.89 & -0.0365 & +4.29 & 69.40 & 85.67 & 94.70 & 98.44 & 0.0372 & 95.27\\
MLP & 0.4345 & 68.18 & 69.33 & 68.75 & +0.0147 & \textbf{\color{ruby}-0.77} & 69.33 & 85.75 & 94.68 & 98.45 & 0.0370 & 95.34\\
RF & 0.4103 & 69.77 & 70.54 & \textbf{70.15} & -0.0178 & \textbf{+1.40} & 70.54 & 87.36 & 95.70 & 98.69 & 0.0336 & 95.53\\\hline
\multicolumn{11}{c}{A positive $\Delta F_1$ score (likewise a negative $\Delta$Brier score) indicates performance relative to ILR from Table~\ref{supp-tab:multi-class-geozone-ilr-performance}}\\
\end{tabular}
}
\end{table}
In Table~\ref{supp-tab:multi-class-geozone-pwlr-performance}, the performance of PWLR is compared with ILR. Please note that $\Delta F_1$ now describes changes relative to ILR rather than the baseline. Previously, the gain was measured against no transformation. For ensemble and tree-based learners (viz. GradBoost, XGB and RF), pairwise log ratio transformation outperforms ILR. This corrobrates previous findings reported by Tolosana-Delgado et al.\,\cite{tolosana2019machine}. The novel observation is that L-SVC, RBF-SVM, MLP, GaussianNB and KNN all perform worse with PWLR in comparison with ILR.

\section{Discussion}\label{sec:supp-summary}
\begin{figure}[!htb]
\centering
\includegraphics[width=160mm]{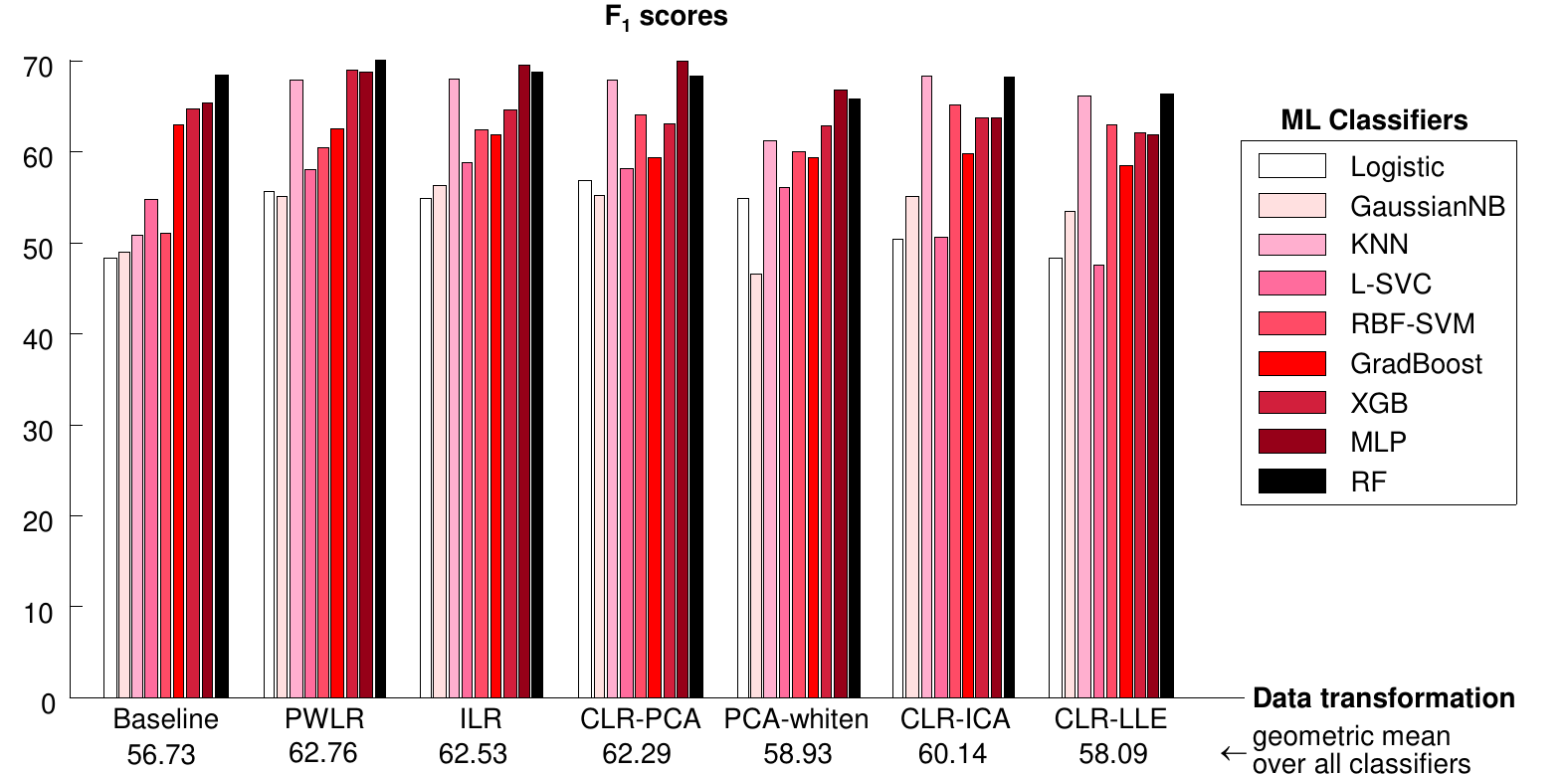}
\caption{Effects of data transformation on geozone classification performance (sorted by transform method)}
\label{fig:effects-of-data-transformation-group-by-xform}
\vspace{\floatsep}
\includegraphics[width=160mm]{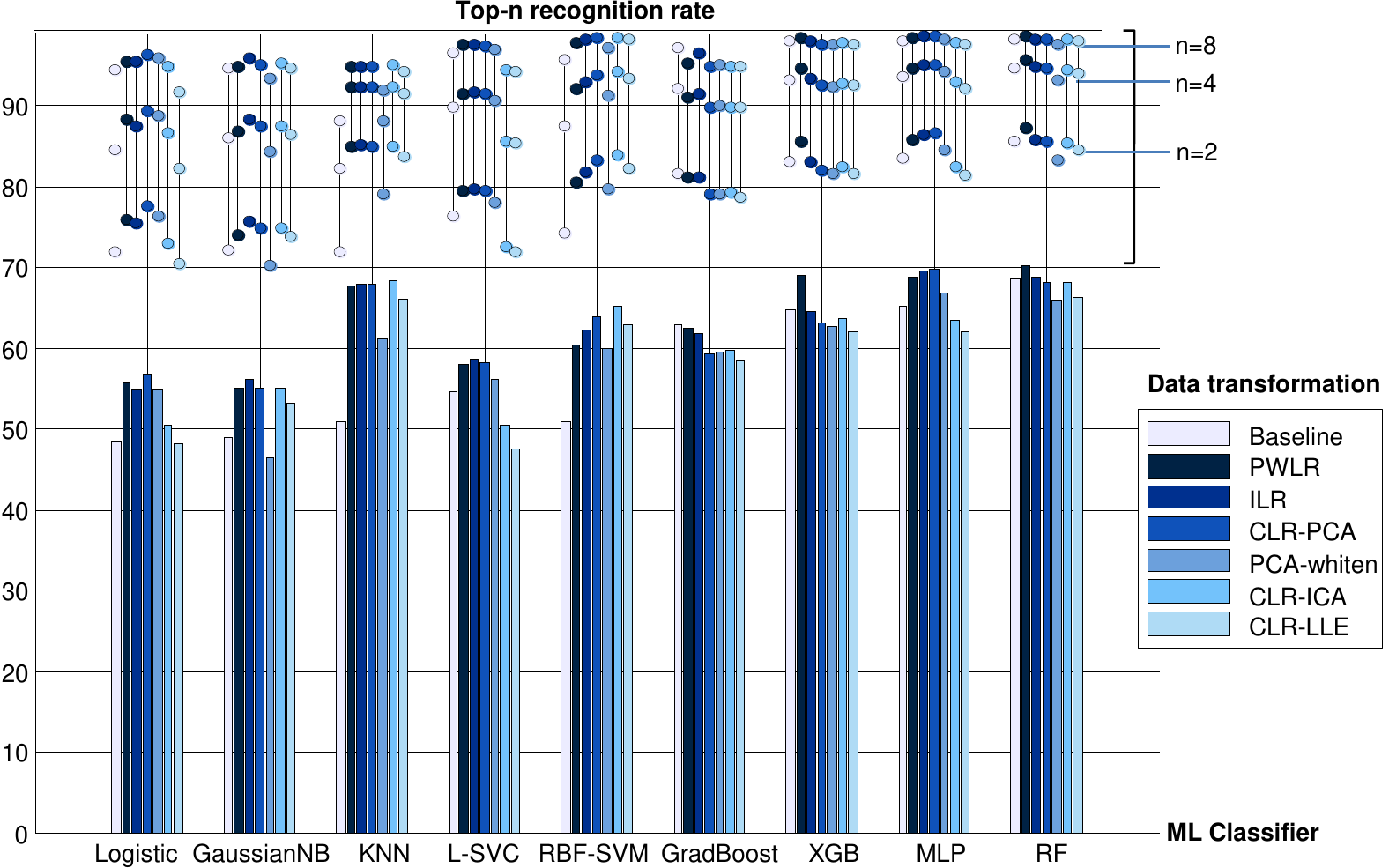}
\caption{Effects of data transformation on geozone classification performance (sorted by classifier)}
\label{fig:effects-of-data-transformation-group-by-classifier}
\end{figure}

The findings (for scenario 1 in particular) can be summarised visually by looking at the utility of each transformation in Fig.~\ref{fig:effects-of-data-transformation-group-by-xform}. In the bar graph, results are grouped by the transformation method along the x-axis, each group in turn assembles the $F_1$ score for the 9 classifiers. To facilitate interpretation, the geometric means of the $F_1$ score computed over all the classifiers are also displayed for each transformation method. As mentioned previously, the isometric log-ratio (ILR) and CLA-PCA generally yield the best results. A notable exception is the case of ensemble and tree-based classifiers where pairwise log-ratio (PWLR) is superior to ILR for GradBoost, XGB and RF.

Figure~\ref{fig:effects-of-data-transformation-group-by-classifier} provides an alternative view of the same results from the perspective of each ML classifier. It illustrates which data transformation each classifier benefits from the most. Drawing attention to each group in the figure from right to left: random forest (RF) and extreme gradient boosting (XGB) are both robust with respect to all data transformations in the sense that their $F_1$ scores show the least variation. Overall, MLP and RF perform the best, particularly when ILR transformation or CLR-PCA is applied.

Between L-SVC and RBF-SVM, it can be seen that CLR-ICA has a deleterious effect on the former and a beneficial effect on the latter. This suggests that both ICA and LLE introduce components (or features) that are not linear separable.

Although the $F_1$ score for KNN appears to be competitive with MLP and RF at first glance, an analysis of the top-$n$ recognition rates at the top of Fig.~\ref{fig:effects-of-data-transformation-group-by-classifier} would reveal its deficiencies. The top-$n$ recognition rate indicates the ability to correctly identify the true geozone in and amongst the top $n$ most likely candidates. In this respect, KNN is inferior to both MLP and RF. The top-8 recall rate for KNN is 94.9\% whereas the corresponding values for MLP and RF are $\ge$98.3\% under ILR. This suggests KNN is less expressive compared with MLP and RF which are able to learn more complex structure within the data and extrapolate patterns on a more consistent basis. KNN incurs a higher false negative rate. When it misdetects a geozone, the likelihood of subsequently finding the correct geozone from its list of next most probable candidates is lower than both MLP and RF.

\newpage
\bibliographystyle{unsrt}  
\bibliography{ms}

\begin{thebibliography}{10}

\bibitem{kucera1998logratio}
Michal Kucera and Bj{\"o}rn~A Malmgren.
\newblock Logratio transformation of compositional data: a resolution of the
  constant sum constraint.
\newblock {\em Marine Micropaleontology}, 34(1-2):117--120, 1998.

\bibitem{verma2006discriminating}
Surendra~P Verma, Mirna Guevara, and Salil Agrawal.
\newblock Discriminating four tectonic settings: Five new geochemical diagrams
  for basic and ultrabasic volcanic rocks based on log-ratio transformation of
  major-element data.
\newblock {\em Journal of Earth System Science}, 115(5):485--528, 2006.

\bibitem{verma2015monte}
Surendra~P Verma.
\newblock Monte carlo comparison of conventional ternary diagrams with new
  log-ratio bivariate diagrams and an example of tectonic discrimination.
\newblock {\em Geochemical Journal}, 49(4):393--412, 2015.

\bibitem{darabi2018evaluation}
F~Darabi-Golestan and A~Hezarkhani.
\newblock Evaluation of elemental mineralization rank using fractal and
  multivariate techniques and improving the performance by log-ratio
  transformation.
\newblock {\em Journal of Geochemical Exploration}, 189:11--24, 2018.

\bibitem{karacan2018mapping}
C~{\"O}zgen Karacan and Ricardo~A Olea.
\newblock Mapping of compositional properties of coal using isometric log-ratio
  transformation and sequential gaussian simulation--a comparative study for
  spatial ultimate analyses data.
\newblock {\em Journal of geochemical exploration}, 186:36--49, 2018.

\bibitem{jaconi2019log}
A~Jaconi, C~Poeplau, L~Ramirez-Lopez, B~Van~Wesemael, and A~Don.
\newblock Log-ratio transformation is the key to determining soil organic
  carbon fractions with near-infrared spectroscopy.
\newblock {\em European Journal of Soil Science}, 70(1):127--139, 2019.

\bibitem{darabi2019multivariate}
Farshad Darabi-Golestan and Ardeshir Hezarkhani.
\newblock Multivariate analysis of log-ratio transformed data and its priority
  in mining science: porphyry and polymetallic vein deposits case studies.
\newblock {\em Maden Tetkik ve Arama Dergisi}, 159(159):185--200, 2019.

\bibitem{andrews2015generalized}
Shawn Andrews and Ghassan Hamarneh.
\newblock The generalized log-ratio transformation: learning shape and
  adjacency priors for simultaneous thigh muscle segmentation.
\newblock {\em IEEE Transactions on Medical Imaging}, 34(9):1773--1787, 2015.

\bibitem{wang2020comparison}
Zong Wang, Wenjiao Shi, Wei Zhou, Xiaoyan Li, and Tianxiang Yue.
\newblock Comparison of additive and isometric log-ratio transformations
  combined with machine learning and regression kriging models for mapping soil
  particle size fractions.
\newblock {\em Geoderma}, 365:114214, 2020.

\bibitem{zhang2020systematic}
Mo~Zhang, Wenjiao Shi, and Ziwei Xu.
\newblock Systematic comparison of five machine-learning models in
  classification and interpolation of soil particle size fractions using
  different transformed data.
\newblock {\em Hydrology and Earth System Sciences}, 24(5):2505--2526, 2020.

\bibitem{tolosana2019machine}
R~Tolosana-Delgado, H~Talebi, M~Khodadadzadeh, and KG~Van~den Boogaart.
\newblock On machine learning algorithms and compositional data.
\newblock In {\em Proceedings of the 8th International Workshop on
  Compositional Data Analysis, Terrassa, Spain}, pages 3--8, 2019.

\bibitem{breiman2001random}
Leo Breiman.
\newblock Random forests.
\newblock {\em Machine Learning}, 45(1):5--32, 2001.

\bibitem{fagerland2008multinomial}
Morten~W Fagerland, David~W Hosmer, and Anna~M Bofin.
\newblock Multinomial goodness-of-fit tests for logistic regression models.
\newblock {\em Statistics in medicine}, 27(21):4238--4253, 2008.

\bibitem{cortes1995support}
Corinna Cortes and Vladimir Vapnik.
\newblock Support-vector networks.
\newblock {\em Machine Learning}, 20(3):273--297, 1995.

\bibitem{anthony2009neural}
Martin Anthony and Peter~L Bartlett.
\newblock {\em Neural network learning: Theoretical foundations}.
\newblock cambridge university press, 2009.

\bibitem{leung2021warpml-arxiv}
Raymond Leung, Mehala Balamurali, and Alexander Lowe.
\newblock Surface warping incorporating machine learning assisted domain
  likelihood estimation: A new paradigm in mine geology modelling and
  automation.
\newblock {\em Mathematical Geosciences (provisionally accepted). Preprint
  available}, ar{X}iv:2103.03923, 2021.

\bibitem{clout2006iron}
JMF Clout.
\newblock Iron formation-hosted iron ores in the {H}amersley {P}rovince of
  {W}estern {A}ustralia.
\newblock {\em Applied Earth Science}, 115(4):115--125, 2006.

\bibitem{leung2020bayesian}
Raymond Leung, Alexander Lowe, Anna Chlingaryan, Arman Melkumyan, and John
  Zigman.
\newblock Bayesian surface warping approach for rectifying geological
  boundaries using displacement likelihood and evidence from geochemical
  assays.
\newblock {\em arXiv e-prints}, arXiv:2005.14427, 2020.

\bibitem{zadrozny2002transforming}
Bianca Zadrozny and Charles Elkan.
\newblock Transforming classifier scores into accurate multiclass probability
  estimates.
\newblock In {\em Proceedings of the ACM SIGKDD International Conference on
  Knowledge Discovery and Data Mining}, pages 694--699. ACM, 2002.

\bibitem{rousseeuw-87}
Peter~J Rousseeuw.
\newblock Silhouettes: a graphical aid to the interpretation and validation of
  cluster analysis.
\newblock {\em Journal of {C}omputational and {A}pplied {M}athematics},
  20:53--65, 1987.

\bibitem{leung2019sample}
Raymond Leung, Mehala Balamurali, and Arman Melkumyan.
\newblock Sample truncation strategies for outlier removal in geochemical data:
  The {MCD} robust distance approach versus t-{SNE} ensemble clustering.
\newblock {\em Mathematical Geosciences}, 53:105--130, 2021.

\bibitem{pedregosa2011scikit}
Fabian Pedregosa, Ga{\"e}l Varoquaux, Alexandre Gramfort, Vincent Michel,
  Bertrand Thirion, Olivier Grisel, Mathieu Blondel, Peter Prettenhofer, Ron
  Weiss, Vincent Dubourg, et~al.
\newblock Scikit-learn: Machine learning in python.
\newblock {\em Journal of Machine Learning Research}, 12(Oct):2825--2830, 2011.

\bibitem{yu2011dual}
Hsiang-Fu Yu, Fang-Lan Huang, and Chih-Jen Lin.
\newblock Dual coordinate descent methods for logistic regression and maximum
  entropy models.
\newblock {\em Machine Learning}, 85(1-2):41--75, 2011.

\bibitem{zhu1997algorithm}
Ciyou Zhu, Richard~H Byrd, Peihuang Lu, and Jorge Nocedal.
\newblock Algorithm 778: {L-BFGS-B}: {F}ortran subroutines for large-scale
  bound-constrained optimization.
\newblock {\em ACM Transactions on Mathematical Software}, 23(4):550--560,
  1997.

\bibitem{murphy2006naive}
Kevin~P Murphy et~al.
\newblock Naive {B}ayes classifiers.
\newblock {\em Lecture {N}otes ({CS}340-{F}all), University of British
  Columbia}, 2006.

\bibitem{song2017efficient}
Yunsheng Song, Jiye Liang, Jing Lu, and Xingwang Zhao.
\newblock An efficient instance selection algorithm for k nearest neighbor
  regression.
\newblock {\em Neurocomputing}, 251:26--34, 2017.

\bibitem{chang2011libsvm}
Chih-Chung Chang and Chih-Jen Lin.
\newblock {LIBSVM}: A library for support vector machines.
\newblock {\em ACM Transactions on Intelligent Systems and Technology (TIST)},
  2(3):27, 2011.

\bibitem{friedman2001greedy}
Jerome~H Friedman.
\newblock Greedy function approximation: a gradient boosting machine.
\newblock {\em Annals of Statistics}, pages 1189--1232, 2001.

\bibitem{chen2016xgboost}
Tianqi Chen and Carlos Guestrin.
\newblock X{GB}oost: A scalable tree boosting system.
\newblock In {\em Proceedings of the 22nd {ACM} {SIGKDD} International
  Conference on Knowledge Discovery and Data Mining}, pages 785--794, 2016.

\bibitem{hinton1990connectionist}
Geoffrey~E Hinton.
\newblock Connectionist learning procedures.
\newblock In {\em Machine Learning}, pages 555--610. Elsevier, 1990.

\bibitem{glorot2010understanding}
Xavier Glorot and Yoshua Bengio.
\newblock Understanding the difficulty of training deep feedforward neural
  networks.
\newblock In {\em Proceedings of the International Conference on Artificial
  Intelligence and Statistics}, pages 249--256, 2010.

\bibitem{he2015delving}
Kaiming He, Xiangyu Zhang, Shaoqing Ren, and Jian Sun.
\newblock Delving deep into rectifiers: Surpassing human-level performance on
  {I}mage{N}et classification.
\newblock In {\em Proceedings of the IEEE International Conference on Computer
  Vision}, pages 1026--1034, 2015.

\bibitem{kingma2014adam}
Diederik~P Kingma and Jimmy Ba.
\newblock Adam: A method for stochastic optimization.
\newblock {\em arXiv e-prints}, arXiv:1412.6980, 2014.

\bibitem{egozcue-03}
J.J. Egozcue, V~Pawlowsky-Glahn, G~Mateu-Figueras, and C~Barcel\'{o}-Vidal.
\newblock Isometric logratio transformations for compositional data analysis.
\newblock {\em Mathematical Geosciences}, 35:279--300, 2003.

\bibitem{tsagris-2011-data}
Michail~T Tsagris, Simon Preston, and Andrew~TA Wood.
\newblock A data-based power transformation for compositional data.
\newblock In J.J. Egozcue, R.~Tolosana-Delgado, and M.I. Ortego, editors, {\em
  4th international workshop on Compositional Data Analysis}, pages 565--572.
  Springer, 2011.

\bibitem{boisvert2009calculating}
J~Boisvert, J~Manchuk, and CV~Deutsch.
\newblock Calculating distance in presence of locally varying anisotropy.
\newblock {\em Mathematical Geosciences}, 41:585--601, 2009.

\bibitem{abdi2010principal}
Herv{\'e} Abdi and Lynne~J Williams.
\newblock Principal component analysis.
\newblock {\em Wiley Interdisciplinary Reviews: Computational Statistics},
  2(4):433--459, 2010.

\bibitem{hyvarinen2000independent}
Aapo Hyv{\"a}rinen and Erkki Oja.
\newblock Independent component analysis: algorithms and applications.
\newblock {\em Neural networks}, 13(4-5):411--430, 2000.

\bibitem{saul2003think}
Lawrence~K Saul and Sam~T Roweis.
\newblock Think globally, fit locally: unsupervised learning of low dimensional
  manifolds.
\newblock {\em Journal of Machine Learning Research}, 4:119--155, 2003.

\bibitem{pawlowsky2006compositional}
Vera Pawlowsky-Glahn and Juan~Jos{\'e} Egozcue.
\newblock Compositional data and their analysis: an introduction.
\newblock {\em Geological Society, London, Special Publications}, 264(1):1--10,
  2006.

\end{thebibliography}

\end{document}